\documentclass[]{hdsr}

\graphicspath{{figs/}}

\usepackage{xcolor,colortbl}

\usepackage{multirow}
\definecolor{Gray}{gray}{0.85}
\definecolor{LightCyan}{rgb}{0.88,1,1}

\newcolumntype{a}{>{\columncolor{Gray}}c}
\newcolumntype{b}{>{\columncolor{white}}c}
\newcolumntype{x}{>{\columncolor{red!20}}c}
\newcolumntype{y}{>{\columncolor{green!50}}c}
\newcolumntype{z}{>{\columncolor{cyan}}c}

\usepackage{endnotes}
\let\footnote=\endnote

\begin{document}

\newgeometry{bottom=1.5in}

\newcommand{\bracketcite}[1]{(\cite{#1})}

\begin{center}

  \title{On the Importance of 3D Surface Information 
for Remote Sensing Classification Tasks}
  \maketitle

  \thispagestyle{empty}
  
  \vspace*{.2in}

\let\cite\textcite

  \begin{tabular}{cc}
    Jan Petrich\upstairs{\affilone,*}, Ryan Sander\upstairs{\affiltwo, \affilthree}, Eliza Bradley\upstairs{\affilone}, Adam Dawood\upstairs{\affilone}, Shawn Hough\upstairs{\affilone}
   \\[0.25ex]
   {\small \upstairs{\affilone} Pennsylvania State University Applied Research Laboratory, United States} \\
   {\small \upstairs{\affiltwo} National Geospatial-Intelligence Agency, United States} \\
   {\small \upstairs{\affilthree} Massachusetts Institute of Technology, United States} \\
  \end{tabular}
  
  \emails{
    \upstairs{*}\href{mailto:jup37@arl.psu.edu}{jup37@arl.psu.edu}
    }
  \vspace*{0.4in}
\newpage
\begin{abstract}
There has been a surge in remote sensing machine learning applications that operate on data from active or passive sensors as well as multi-sensor combinations \bracketcite{MA2019166}. Despite this surge, however, there has been relatively little study on the comparative value of 3D surface information for machine learning classification tasks. Adding 3D surface information to RGB imagery can provide crucial geometric information for semantic classes such as buildings, and can thus improve out-of-sample predictive performance. In this paper, we examine in-sample and out-of-sample classification performance of Fully Convolutional Neural Networks (FCNNs) and Support Vector Machines (SVMs) trained with and without 3D normalized digital surface model (nDSM) information. We assess classification performance using multispectral imagery from the International Society for Photogrammetry and Remote Sensing (ISPRS) 2D Semantic Labeling contest and the United States Special Operations Command (USSOCOM) Urban 3D Challenge. We find that providing RGB classifiers with additional 3D nDSM information results in little increase in in-sample classification performance, suggesting that spectral information alone may be sufficient for the given classification tasks. However, we observe that providing these RGB classifiers with additional nDSM information leads to significant gains in out-of-sample predictive performance. Specifically, we observe an average improvement in out-of-sample all-class accuracy of 14.4\% on the ISPRS dataset and an average improvement in out-of-sample F1 score of 8.6\% on the USSOCOM dataset. In addition, the experiments establish that nDSM information is critical in machine learning and classification settings that face training sample scarcity.
\end{abstract}
\end{center}

\vspace*{0.15in}
\hspace{10pt}
  \small	
  \textbf{\textit{Keywords: }} {Remote Sensing, 3D, Deep Learning, Classification, Data Fusion}

\section{Introduction}
\label{sec1}

Significant resources are required to establish robust classification performance for remote sensing applications which can range from composite distributed land uses, for example urban versus rural development, to feature-specific mapping, such as roads, building and vehicles. Technical challenges naturally arise from the sheer variety of appearances and viewing angles for objects of interest, as well as the lack of annotations or labels relative to the overall volume of data. \\

Machine learning and artificial intelligence have made great strides in recent years due to the advent of hardware accelerated computing and the development of symbolic math libraries. However, a unified framework for remote sensing classification applications is still beyond reach \bracketcite{ball2017comprehensive}\bracketcite{zhu2017deep}. The diversity of data collection methods, e.g. space-based or airborne, and sensor modalities, e.g. Lidar, RGB, and hyperspectral imagery, necessitate continuous adaptation of the underlying classification framework based on classification task, desired accuracy, computational bandwidth, and available datasets. \\

A range of machine learning methodologies are available for executing object classification from remotely sensed data including convolutional neural networks (CNNs) \bracketcite{maturana20153d}\bracketcite{zhao2018classifying}, decision trees \bracketcite{blaha2016large}, and Support Vector Machines (SVMs) \bracketcite{lai20093d} \bracketcite{wang2007support}. Although inherently different, all of those methods have been shown to be effective and successful given specific datasets and class labels for training and cross-validation. Interestingly, many algorithms originate from outside the remote sensing community, e.g. from biomedical image segmentation \bracketcite{ronneberger2015u}, and have been modified to ingest mission-specific sensor modalities in order to classify different objects of interest. \\

The feasibility of applying machine learning concepts has been further studied, and showcased through various challenges, such as the USSOCOM Urban 3D challenge \bracketcite{Urban3D2017}\bracketcite{goldberg2018urban} or the ISPRS 2D and 3D Semantic Labeling contests \bracketcite{inproceedings}. Several submitted algorithms yielded high accuracy, even without using 3D surface information \bracketcite{audebert2018beyond}. However, given the recent success, it remains important to formally assess information content of available sensor modalities, and to infer generalizability of the resulting classifier if presented with inherently different, previously unseen test samples. In other words, given a specific object recognition task, we seek to answer the following questions. First, what are the preferred sensor modalities given the quantity and quality of training samples? After all, adding sensor modalities usually incurs cost and logistical considerations. Second, is the resulting classifier useful when applied outside the location or domain in which training data was collected? \\

This paper expands upon the findings in \bracketcite{8519225}, in which the authors acknowledge that the majority of ongoing work is focused on developing classification strategies with little regard to the information value provided by the sensor modality of the input data. In \bracketcite{8519225}, multi-spectral imagery is fused with DSM information in order to study the impact on the resulting classification performance. However, DSM information is pre-processed to extract specific features based on the 3D structure tensor \bracketcite{weinmann2016reconstruction} such as linearity, planarity, and sphericity, as well as the Normalized Difference Vegetation Index (NDVI) from \bracketcite{tucker2001higher}. This paper expands upon the work in \bracketcite{8519225} by considering multiple datasets and by examining training and test performance separately. Additionally, cross-city validation is added to formally assess classification performance when extrapolating the classifier from one city to another. \\

To partially close this gap between formally assessing information content and achieving high classification performance, this paper examines the importance of 3D surface information when generalizing trained classifiers using two distinct machine learning frameworks, and two different publicly available datasets as test cases. In order to derive results that are agnostic to the chosen classification methodology, we specifically select one classification scheme of high complexity, i.e. a Fully Convolutional Neural Network (FCNN) architecture, and one classification scheme of low complexity, i.e. a Support Vector Machine (SVM).\\

The novelties and technical contributions of this paper are two-fold. First, we examine generalizability of classification frameworks when trained and tested on different geographic locations by formally evaluating out-of-sample performance of the classifier when trained with and without 3D DSM information. Different geographic locations often imply deviations in the underlying urban infrastructure, architecture, and thus the appearance and structure of objects of interest such as buildings and roads. Second, we assess the same generalizability when facing scarce training data. Therefore, we will formally evaluate out-of-sample performance when trained with and without 3D DSM information while the number of training samples is gradually reduced. \\

As acknowledged in \bracketcite{chen2018residual}, the majority of research focuses on improving classification tasks, while   ``only little attention has been paid to the input data itself". Specifically, information content is often overlooked in light of achievable performance metrics for object classification and segmentation. In addition, the work in \bracketcite{chen2018residual} explores the utility of handcrafted, geometric features over machine-derived ones. The information content of specific feature types is studied in \bracketcite{gevaert2016classification}, in which the authors distinguish between point-based and segment-based features. Dependent on the available sensor modalities, only some feature types may be available in any given scenario. Their relative importance for object classification can then be formally assessed using an SVM classifier for different scenarios and different combinations of feature types. Although classification is tested via cross-validation, training and testing samples are extracted from the same geographical location, which may limit the diversity of building footprints the classifier is exposed to. This paper follows a similar approach by formally evaluating not only classification performance, but also generalizability using two distinct classification frameworks. Therefore, we attempt to establish the notion of information content independent of the underling classification strategy. In addition, our study focuses on entire sensor modalities instead of individual features that are derived from those modalities.\\

This paper is organized as follows: Section \ref{sec:TechnicalApproach} outlines the technical approach. Experimental validation is presented in Section \ref{experimental_validation}, while results are split into cross-city validation in Section \ref{cross_city_validation} and validation for varying training sample proportions in Section \ref{small_sample_validation}. Section \ref{conclusion} concludes the study.
\section{Technical Approach} 
\label{sec:TechnicalApproach}
This section describes the underlying classification approach and performance assessment. Section \ref{sec:BinaryandMultiClassClassification} provides an overview of binary and multi-class classification in a semantic segmentation setting. Section \ref{imbalance_correction} discusses class balancing for imbalanced classification tasks. Section \ref{ClassificationArchitectures} outlines the two classification architectures used for this study, i.e. a Support Vector Machine (SVM) and a Fully Convolutional Neural Network (FCNN). The two publicly available datasets, including objects of interest used for training and testing, are described in Section \ref{DatasetsandObjectsofInterest}. Section \ref{ValidationScenarios} summarizes the performance assessment strategies as well as the two different validation settings: (1) cross-city validation and (2) reduction of sample proportion for training.

\subsection{Binary and Multi-Class Classification}
\label{sec:BinaryandMultiClassClassification}
This work aims to (i) develop robust binary and multi-class semantic segmentation classifiers for remote sensing, and (ii) test the overall generalizability of these classifiers as a function of sensor modality. Semantic segmentation can be conceptualized as a classification problem in which we predict a class label for each pixel. We consider each of these pixel-level classification problems in binary and multi-class classification settings \bracketcite{janocha2017loss}. One common optimization objective for finding optimal parameter sets in binary and multi-class classification tasks is two-dimensional cross-entropy, which we utilize in this work for our FCNN models. Below, we discuss the objective functions for both of these classification schemes.
\subsubsection{Binary Classification}
\label{binaryclassification}
For binary classification, the classifier learns by minimizing a loss function over a set of parameters $\boldsymbol{\theta}_b^*$, in this case pixel-wise Negative Log Likelihood (NLL) with a single class.
\begin{equation}
    \boldsymbol{\theta}_b^* = \arg\min_{\;\boldsymbol{\theta} \in \Theta}\;\left(-\sum_{i=1}^{m}\sum_{j=1}^{n}Y_{i,j}\log(P(X; \boldsymbol{\theta})_{i,j})+(1-Y_{i,j})\;\log(1-P(X; \boldsymbol{\theta})_{i,j})\right)
\end{equation}
Where $Y_{i,j} \in \{0, 1\}$ and $P(X;\boldsymbol{\theta})_{i,j} \in [0, 1]$ denote the ground truth and predicted labels, respectively, at pixel $(i, j)$ for each training image tile, and $X$ denotes the input features (1, 3, or 4-channel image). Both $X$ and $Y$ are indexed as 2D arrays with (height, width) given as $(m, n)$. This classification methodology is utilized for the USSOCOM Urban3D dataset for building footprint classification.
\subsubsection{Multiclass Classification}
\label{multiclassclassication}
A multiclass classifier minimizes a similar objective to find parameters $\boldsymbol{\theta}_m^*$, except rather than considering a single class, e.g. a building, this classifier makes pixel-wise predictions over multiple classes. This classifier learns by minimizing the following objective, which corresponds to pixel-wise Negative Log Likelihood (NLL) over $N$ classes.
\begin{equation}
    \boldsymbol{\theta}_m^* = \arg\min_{\;\boldsymbol{\theta} \in \Theta}\;\left(-\sum_{i=1}^{m}\sum_{j=1}^{n}\sum_{k=1}^{N}Y_{i,j,k}\log(P(X; \boldsymbol{\theta})_{i,j,k})\right)
\end{equation}
Where $Y_{i,j,k} \in \{0, 1\}$ and $P(X, \boldsymbol{\theta})_{i,j,k} \in [0, 1]$ denote the ground truth and predicted labels, respectively, at pixel $(i, j)$ for class $k \in \{0, 1, \;\cdots\;, N\}$ for each training tile $X$, which is indexed as a 2D array of shape $(m, n)$ as above. \\ \\
To make optimal semantic segmentation predictions, we use gradient descent methods \bracketcite{ruder2016overview} to iteratively solve for a set of weights $\boldsymbol{\theta}$ that minimize the objectives presented above.
\subsection{Class Imbalance Correction via Class Weighting}
\label{imbalance_correction}
In addition to computing element-wise cross-entropy loss for each pixel in the image, we weight each pixel using its inverse class frequency, i.e. if class $i$ occurs in a dataset with fraction $f_i \in (0, 1]$, then our weighting corresponds to the inverse of this fraction: $w_i= \frac{1}{f_i}$. This weighting scheme is used to correct for class imbalance in our datasets, and corresponds to a weighted Negative Log Likelihood (wNLL) or weighted Cross-Entropy objective \bracketcite{aurelio2019learning}. Using a weighted cost function allows for learning a classifier that is not biased toward the most common class. With the incorporation of these weights, our objectives become: \\ \\
\textbf{Weighted Binary Classification:}
\begin{equation}
    \boldsymbol{\theta}_{b,w}^* = \arg\min_{\;\boldsymbol{\theta} \in \Theta}\;\left(-\sum_{i=1}^{m}\sum_{j=1}^{n}w(Y_{i,j})\left(Y_{i,j}\log(P(X;\boldsymbol{\theta})_{i,j})+(1-Y_{i,j})\;\log(1-P(X; \boldsymbol{\theta})_{i,j})\right)\right)
\end{equation}
Where $w(Y_{i,j}) \in \mathbb{R}^{+}$ denotes the weight assigned to pixel $(i,j)$ based off of its ground truth class, $Y_{i,j}$. Note that $Y_{i,j} \in \{0, 1\}.$\\ \\
\textbf{Weighted Multi-Class Classification:}
\begin{equation}
    \boldsymbol{\theta}_{m,w}^* = \arg\min_{\;\boldsymbol{\theta} \in \Theta}\;\left(-\sum_{i=1}^{m}\sum_{j=1}^{n}\sum_{k=1}^{N}w(c_{i,j})Y_{i,j,k}\log(P(X; \boldsymbol{\theta})_{i,j,k})\right)
\end{equation}
Where $w(c_{i,j}) \in \mathbb{R}^{+}$ denotes the weight assigned to pixel $(i,j)$ based off of its ground truth class $c_{i,j}$. Note that $c_{i,j} \in \{0, 1, \;\cdots\;, N\}$ denotes the class, rather than $Y_{i,j,k} \in \{0, 1\}.$ \\

Our FCNN classifiers, SegNet and SegNet Lite, which are described in detail below, each leverage these weighted objectives, and our classification results are reported as balanced/class frequency adjusted.
\subsection{Classification Architectures}
\label{ClassificationArchitectures}
To fully evaluate the importance of 3D nDSM information independent of the underlying classification algorithm, two distinct classification architectures have been selected for experimentation. First, we consider a relatively simple, well-known SVM framework. Second, we apply an FCNN architecture. These two classifers are outlined in detail below. \\

\subsubsection{Architecture 1 – Support Vector Machine (SVM)} \label{svm_background} 
\begin{figure}[ht]
    \centering
    \includegraphics[width=10cm]{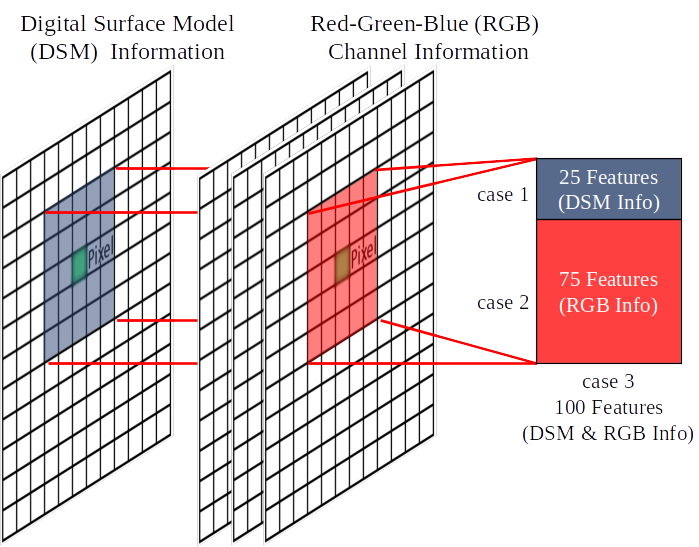}
    \caption{Feature extraction for SVM classification using a 5x5 neighborhood and a 1 channel (DSM only), 3 channel (RGB-only), or 4 channel (DSM \& RGB) representation.}
    \label{svm_diagram}
\end{figure}

SVM is a discriminative classifier that aims to separate sets of feature vectors based on the assigned labels by finding the maximum margin hyperplane \bracketcite{cortes1995support}. Separation can be accomplished through linear hyperplanes or nonlinear hypersurfaces, which are constructed in a feature space through numerical optimization with kernel functions. Figure \ref{svm_diagram} illustrates the application of a $5 \times 5$ neighborhood in the imagery to construct $n$-dimensional feature vectors $f_i \in \mathbb{R}^n$, one for each pixel (or data sample) $i$. For case 1 (one channel), we extract features only from the DSM information to obtain a feature vector $f_{1, i} \in \mathbb{R}^{25}$. Similarly, for case 2 (three channels), we utilize only RGB information to obtain $f_{2, i} \in \mathbb{R}^{75}$. For case 3 (four channels), we find $f_{3, i} \in \mathbb{R}^{100}$ by concatenating $f_{1, i}$ and $f_{2, i}$, thereby constructing the full feature vector fusing DSM and RGB information. Therefore, classification will be carried out in 25, 75 or 100-dimensional feature spaces.\\ 

Based on ground truth information and training data available, labels $y_i \in L$ can be assigned to each feature vector $f_i$. Here, $L$ denotes the set of all labels with cardinality $|L| = N$, where $N$ is the number of unique labels contained in the dataset. Once feature vectors have been constructed and labels have been assigned, supervised training via SVM can be accomplished using any available machine learning library. For this study, nonlinear hypersurfaces are constructed using a Gaussian/Radial Basis Function (RBF) kernel, implemented with MATLAB’s \texttt{fitcecoc} module \bracketcite{fitcecoc}. \\

\subsubsection{Architecture 2 – Fully Convolutional Neural Network (FCNN)}  \label{segnet_background}  
\begin{figure}[ht]
    \centering
    \includegraphics[width=12cm]{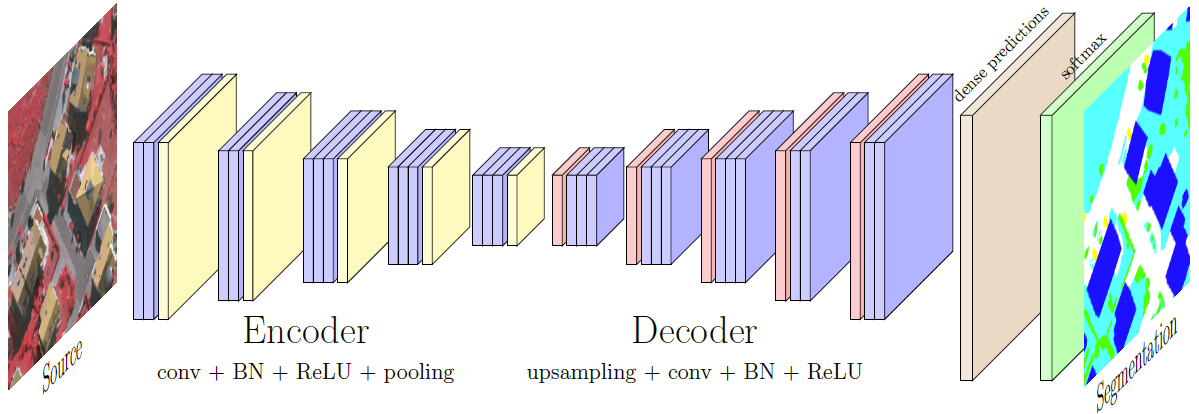}
    \caption{SegNet architecture from \bracketcite{audebert2018beyond}\bracketcite{badrinarayanan2017segnet} utilizing a deep encoder-decoder structure for image segmentation and object classification. To perform nonlinear up-sampling, the SegNet decoder leverages pooling indices computed in the encoder layers of the network and connected from the encoder to the decoder via skip connections \bracketcite{badrinarayanan2017segnet}.}
    \label{segnet_diagram}
\end{figure}

Although originally developed for image segmentation tasks, the Segmentation Network, or SegNet, architecture presented in \bracketcite{badrinarayanan2017segnet} and depicted in Figure \ref{segnet_diagram} has recently gained increased attention for object classification in remote sensing applications \bracketcite{audebert2018beyond}. SegNet is characterized by a Fully Convolutional Neural Network (FCNN) architecture with an encoder-decoder structure. Image information is initially down-sampled throughout the five encoder blocks (left side of Figure \ref{segnet_diagram}) using convolution operations, batch normalization, nonlinear activation functions, and pooling. These encoding blocks create a latent representation of the input image, characterized by spatial features extracted from the convolutional encoding layers. Throughout the five decoding blocks (right side of Figure \ref{segnet_diagram}), segmentation information is reconstructed from this latent representation to the full image resolution using convolution operations, batch normalization, nonlinear activation functions, and nonlinear up-sampling blocks. To perform nonlinear up-sampling, the SegNet decoder leverages pooling indices computed in the encoder layers of the network and connected from the encoder to the decoder via skip connections \bracketcite{badrinarayanan2017segnet}. The input layer can be modified to ingest one channel (DSM only), three channel (RGB-only), or four channel (DSM \& RGB) data.\\

\begin{figure}[ht]
    \centering
    \includegraphics[width=15cm]{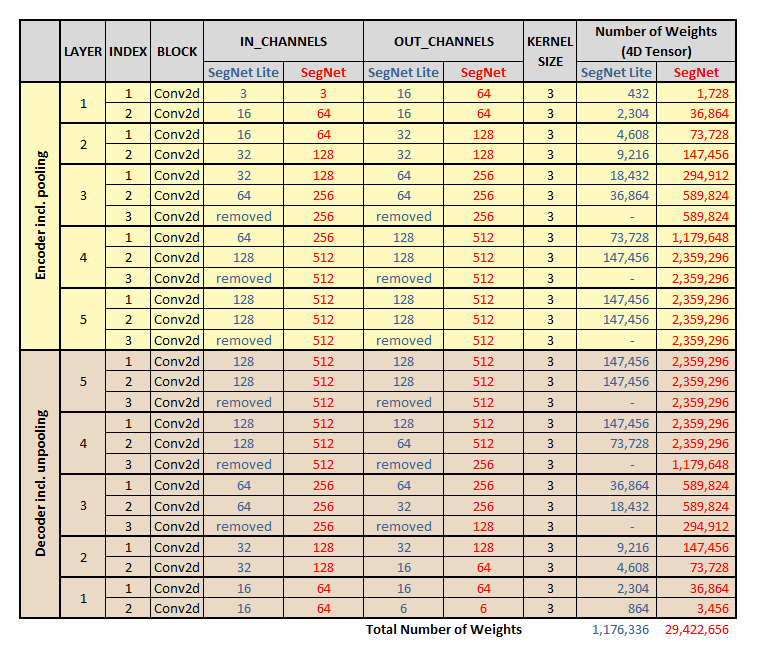}
    \caption{Comparison of \texttt{Segnet} and \texttt{Segnet Lite} architectures by number of indices per layer, number of input and output channels, weights per layer, and total weights. Note that the Segnet Lite architecture limits the number of layers per block to two and reduces the output channels for each layer by 75 percent.}
    \label{segnet_comparison}
\end{figure}

Each of the five encoder blocks and five decoder blocks consists of two to three convolutional layers. For an input window size of $256 \times 256$ pixels, it can be shown that the original SegNet structure from \bracketcite{badrinarayanan2017segnet} consists of roughly 30 million weights. By limiting the number of convolutional layers per block to two and reducing the output dimensions (or channels) for each layer by 75 percent, we construct a similar, yet lightweight SegNet architecture consisting of only 1.2 million weights, reducing the total number of weights by 96\%. For experiments carried out in Section \ref{experimental_validation}, we will refer to the original SegNet architecture as \textit{SegNet} (\texttt{SegNet}) and the lightweight SegNet structure with a reduced number of weights as \textit{SegNet Lite} (\texttt{SegNet Lite}). Figure \ref{segnet_comparison} compares our \texttt{SegNet} (red) and \texttt{SegNet Lite} (blue) architectures, listing the number of weights per layer and the total number of weights. Table \ref{tab:nn_comparison} provides a consolidated view of these two architectures. \\

\begin{table}[ht]
    \centering
    \begin{tabular}{|c|c|c|c|}
        \hline
        \rowcolor{Gray}
        \textbf{Neural Architecture} & \textbf{Total Parameters} & \textbf{Channels (Relative to SegNet)} & \textbf{Kernel Size} \\
        \hline
        \texttt{SegNet} & 29,422,656 & 1.0x & 3\\
        \hline
        \texttt{SegNet Lite} & 1,176,336 & 0.25x & 3\\
        \hline
    \end{tabular}
    \caption{Comparative summary between our two Fully Convolutional Neural Network architectures, \texttt{SegNet} and \texttt{SegNet Lite}. These metrics are based off of Figure \ref{segnet_comparison}.}
    \label{tab:nn_comparison}
\end{table}

Both of these SegNet neural architecture variants were implemented using PyTorch \bracketcite{NEURIPS2019_9015}, which supports streamlined GPU acceleration. Source code was provided though a git repository\footnote{\href{https://github.com/nshaud/DeepNetsForEO}{https://github.com/nshaud/DeepNetsForEO}} shared by the authors of \bracketcite{audebert2018beyond}. Some modifications were made (i) to ingest composite images combining spectral and DSM information, and (ii) to transform the SegNet (\texttt{SegNet}) architecture into the SegNet Lite (\texttt{SegNet Lite}) architecture. 

\subsection{Datasets and Objects of Interest}
\label{DatasetsandObjectsofInterest}
This section describes the two datasets used for performance assessment, including objects of interest. Although there exists a wide variety of available datasets for examining remote sensing applications, this work focuses on (i) high-resolution satellite imagery provided through the USSOCOM Urban 3D Challenge \bracketcite{Urban3D2017}\bracketcite{goldberg2018urban} and (ii) aerial imagery released by the International Society for Photogrammetry and Remote Sensing (ISPRS) in support of their 2D Semantic Labeling Contest \bracketcite{inproceedings}. In this paper, we refer to these datasets by the originator (e.g. USSOCOM for the Urban 3D Challenge and ISPRS for the ISPRS 2D Semantic Labeling Contest). To establish the core results of this study, we apply both the SVM and SegNet classifiers to the ISPRS and USSOCOM datasets. \\

\subsubsection{USSOCOM Urban3D Dataset}
\begin{figure}[ht]
    \centering
    \includegraphics[width=15cm]{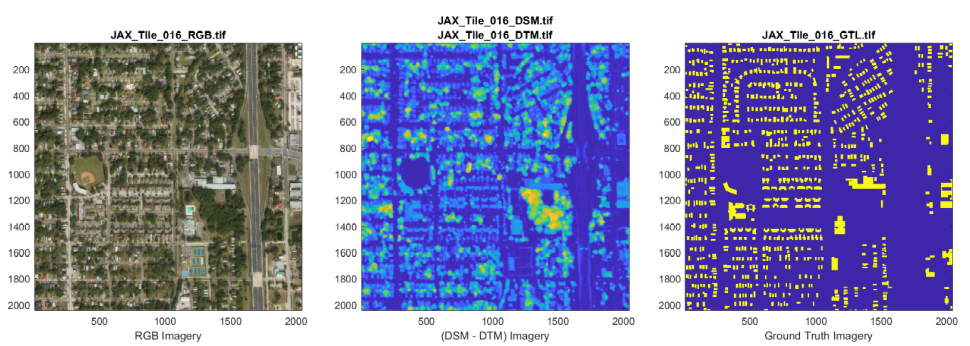}
    \caption{Sample tile from USSOCOM Urban 3D Challenge dataset for Jacksonville, FL showing RGB imagery (left), nDSM info (center), and annotated ground truth for building footprints (right).}
    \label{jax_tiles}
\end{figure}

\begin{figure}[ht]
    \centering
    \includegraphics[width=6cm]{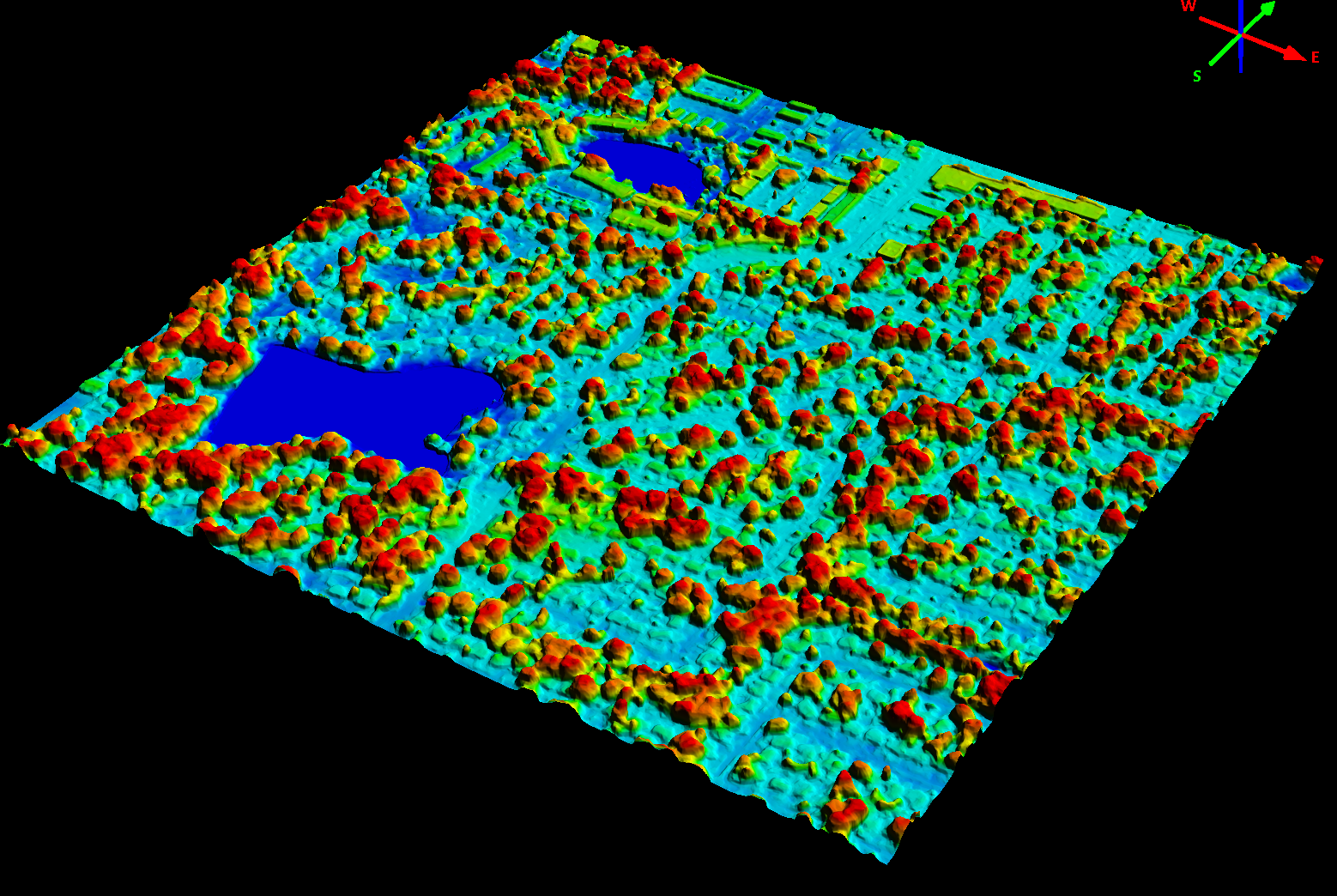}
    \caption{Another view of a sample nDSM (Jacksonville Tile 23) from the USSOCOM dataset.}
    \label{jax_dsm}
\end{figure}

The USSOCOM dataset contains orthorectified red-green-blue (RGB) imagery of three US cities: (i) Jacksonville, FL, (ii) Tampa, FL, and (iii) Richmond, VA with a resolution of 50-centimeters ground sample distance (GSD). The data was collected via commercial satellites and additionally provides coincident 3D Digital Surface Models (DSM) as well as Digital Terrain Models (DTM). These DSMs and DTMs are derived from multi-view, EO satellite imagery at 50-centimeter resolution, rather than through LiDAR (Light Detection and Ranging) sensors. DSM and DTM information is used to generate normalized DSM (nDSM) information, i.e. $\text{nDSM} = \text{DSM} - \text{DTM}$. All imagery products were created using the Vricon (now Maxar Technologies) production pipeline using 50-centimeter DigitalGlobe satellite imagery. Buildings are the only objects of interest, i.e. $L = \{0, 1\}$ and $|L| = 2$, with roughly 157,000 annotated building footprints contained in the data. These ground truth building footprints are generated through the use of a semi-automated feature extraction tool in tandem with the HSIP 133 cities dataset \bracketcite{Urban3D2017}. Figure \ref{jax_tiles} shows one of the 144 Jacksonville tiles from the USSOCOM dataset with RGB imagery on the left, nDSM information (i.e. the difference between surface and terrain) in the center, and annotated ground truth for building footprints on the right (buildings in yellow with background in blue). Tiles are $2048 \times 2048$ pixels and span an area of roughly 1km$^2$. Figure \ref{jax_dsm} depicts an nDSM of one of the Jacksonville tiles from the USSOCOM dataset \bracketcite{Urban3D2017}\bracketcite{goldberg2018urban} shown from a different point of view. More information about the USSOCOM Urban3D dataset can be found in \bracketcite{Urban3D2017} and \bracketcite{goldberg2018urban}. \\

\subsubsection{ISPRS Semantic Labeling Dataset}
\begin{figure}[ht]
    \centering
    \includegraphics[width=14cm]{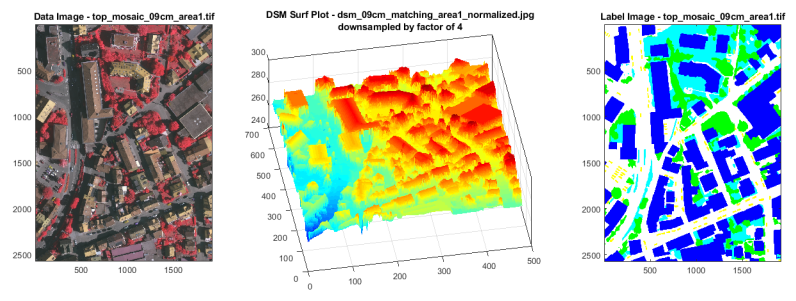}
    \caption{Sample tile from the ISPRS dataset for Vaihingen, Germany showing IRRG imagery (left), nDSM information (center), and color-coded ground truth for six object classes of interest (right).}
    \label{isprs_sample}
\end{figure}

The ISPRS dataset contains infrared, red, and green (IRRG) bands for two locations: (i) Vaihingen, Germany, and (ii) Potsdam, Germany. The GSDs are 9 centimeters and 5 centimeters, respectively. DSM and nDSM information is generated via dense image matching using Trimble INPHO 5.3 software. In order to avoid areas without data (`holes') in the True Orthophoto (TOP) and DSM, dataset patches were selected only from the central region of the TOP mosaic, i.e. not at the boundaries. Any remaining (very small) holes in the TOP and the DSM were interpolated. nDSM imagery is produced using a fully automatic filtering workflow without any manual quality control. 32-bit grey levels are used to encode heights for the DSMs and nDSMs in the TIFF format \bracketcite{inproceedings}. Ground truth annotations for objects of interest are provided for $|L| = 6$  classes, i.e. impervious surfaces (i.e. roads), buildings, low vegetation, trees, cars, and clutter. Figure \ref{isprs_sample} presents a sample of the ISPRS dataset with IRRG imagery on the left, nDSM information in the center, and ground truth information on the right. Ground truth is color-coded for roads (white), buildings (blue), low vegetation (cyan), trees (green), cars (yellow) and clutter (red). \\

For nDSM imagery produced in both the ISPRS and USSOCOM datasets, it is unclear if the authors leveraged any techniques to mitigate occlusion due to objects such as trees, which can have a substantial, seasonally-dependent effect on the ground-truth accuracy of the semantic labels \bracketcite{park2019creating} present in both of these datasets. For the USSOCOM dataset, the authors also note that producing DTM imagery products from overhead imagery remains an open research question \bracketcite{Urban3D2017}.

\subsection{Validation Settings}
\label{ValidationScenarios}
This study provides experimental validation for two scenarios demonstrating the importance of 3D surface information for remote sensing classification tasks. \\

First, we seek to establish performance metrics for cross-city validation when using classifiers that were trained with and without nDSM information. Previous work in this domain concluded that in-sample performance drops only slightly when depriving the classifier (Segnet in \bracketcite{audebert2018beyond}) of nDSM information. However, the impact of nDSM information on out-of-sample performance, i.e. cross-city performance, has not been studied and formally assessed yet, and is one of the major contributions of this work. \\

In addition to cross-city validation, we study the impact of nDSM information when training the classifier using scarce data. Therefore, we will reduce the number of training samples while assessing overall out-of-sample performance of the resulting classifier when trained both with and without nDSM information. \\

Table \ref{ussocom_training} summarizes the cross-city training and testing methodologies for the USSOCOM dataset, while Table \ref{isprs_training} summarizes the cross-city training and testing methodologies for the ISPRS dataset. As noted, we establish the core results of this study by evaluating the importance of DSM information for classification. For this, we use both the SVM and SegNet classifiers for both the USSOCOM and ISPRS datasets. \\

\begin{table}[ht]
    \centering
    \caption{USSOCOM training and testing (in-sample and out-of-sample) procedures for SVM and SegNet. For evaluating the SegNet classifier on the USSOCOM dataset, we only test out-of-sample performance.}
    \begin{tabular}{|c|c|c|c|}
    \hline
    \rowcolor{Gray}
     & \multicolumn{3}{|c|}{\textbf{Classifier Architecture}} \\
    \hline 
    \rowcolor{Gray}
    \textbf{Type of Dataset} & \textbf{SVM} & \multicolumn{2}{|c|}{\textbf{SegNet}}  \\
    \hline
    Training & Jacksonville, FL & Tampa, FL & Tampa, FL  \\
    \hline
    In-Sample Testing & Jacksonville, FL & - & -  \\
    \hline
    Out-of-Sample Testing & Tampa, FL & Richmond, VA & Jacksonville, FL  \\
    \hline 
    \end{tabular}
    \label{ussocom_training}
\end{table}
\begin{table}[ht]
    \centering
    \caption{ISPRS training and testing (in-sample and out-of-sample) procedures for our classification architectures: SVM, SegNet Lite, and SegNet.}
    \begin{tabular}{|c|c|c|c|}
    \hline
    \rowcolor{Gray} 
     & \multicolumn{3}{|c|}{\textbf{Classifier Architecture}} \\
    \hline 
    \rowcolor{Gray}
    \textbf{Type of Dataset} & \textbf{SVM} & \textbf{SegNet Lite} & \textbf{SegNet}  \\
    \hline
    Training & Vaihingen tiles 1-12  & Vaihingen tiles 1-12 & Vaihingen tiles 1-12 \\
    \hline
    In-Sample Testing & Vaihingen tiles 13-16  & Vaihingen tiles 13-16 & Vaihingen tiles 13-16 \\
    \hline
    Out-of-Sample Testing & Potsdam$^*$ & Potsdam$^*$ & Potsdam$^*$ \\
    \hline 
    \end{tabular}
    \label{isprs_training}
\end{table}
*It is important to note that the ISPRS – Potsdam data (5 centimeters GSD) will be down-sampled by a ratio of 9:5 to achieve a GSD of 9 centimeters, which is used for training using the ISPRS - Vaihingen data. \\

In addition to what is summarized in Tables \ref{ussocom_training} and \ref{isprs_training}, these training and evaluation procedures also allow for formal comparisons between the SVM and SegNet classifiers on both the USSOCOM and ISPRS datasets.
\section{Experimental Validation}
\label{experimental_validation}
This section summarizes the experimental validation including performance assessment results. The section is divided with Section \ref{cross_city_validation} addressing cross-city validation and Section \ref{small_sample_validation} describing the impact of reducing available training samples. Classification performance with and without DSM information is the main focus of this study. 
\subsection{Cross-City Validation}
\label{cross_city_validation}
Our first experimental study applies SVM and SegNet classifiers to the ISPRS dataset. Specifically, both networks were trained using three cases: (i) IRRG information only, (ii) nDSM information only, and (iii) nDSM \& IRRG information combined. Training was conducted using 12 out of 16 tiles from the ISPRS Vaihingen dataset. Training times were approximately 20 hours for the SegNet and 3 hours and 20 minutes for SegNet Lite. \\

In our second experimental study, we apply SVM and SegNet classifiers to the USSOCOM dataset. For SVM, we first train the SVM classifier from Section \ref{svm_background} on the USSOCOM dataset using all of the available 144 Jacksonville tiles. Training samples were down-selected randomly by a factor of 1,000 in order to circumvent memory limitations, reduce the number of resulting support vectors, and to allow for adequate training and inference times. SVM training then yields three classifiers, one for training with RGB \& nDSM information, one for training with RGB information only, and one for training with nDSM information only. For SegNet, we follow a similar training procedure: we train the SegNet classifier from Section \ref{segnet_background} on all 144 available Tampa tiles. Down-selection is not performed for the SegNet classifier, i.e. the classifier is trained on all pixels on all training tiles. SegNet training similarly yields three classifiers, one for training with RGB \& nDSM information, one for training with RGB information only, and one for training with nDSM information only. \\

Results from these two aforementioned experimental studies are discussed below.
\subsubsection{ISPRS Dataset Results}
\label{isprs_dataset_results}
The results outlined in Tables \ref{isprs_results_full_segnet} and \ref{isprs_results_segnet_lite} replicate the findings from \bracketcite{audebert2018beyond}. These tables summarize resulting classification accuracy for the SegNet and SegNet Lite architectures, respectively. Individual rows are associated with different objects of interest, while columns cite the corresponding in-sample (Vaihingen) and out-of-sample (Potsdam) class-balanced accuracies for the three training cases. Note that random guessing would yield a total accuracy of $1:6 \approx 0.1667$ for a six class classification problem. Our results indicate that for SegNet, in-sample classification performance is not impacted significantly when depriving the classifier of nDSM information. In fact, accuracy drops less than 0.5\% for either SegNet classifier between the nDSM \& IRRG case and the IRRG-only case. For SVM, we observe a more significant drop in in-sample accuracy of 9\% between the nDSM \& IRRG case and the IRRG-only case. However, unlike in-sample performance, total accuracy drops for out-of-sample validation by 25\%, from 65\% to 40\% for the SegNet, by 5\%, from 50\% to 45\% for SegNet Lite, and 13\%, from 54\% to 41\% for SVM, when excluding nDSM information from training. Performance losses are noticeable across all objects of interest. Although the nDSM-only classifier performs worst in-sample, for SegNet, it outperforms the IIRG-only classifier by 8\% out-of-sample, and for SegNet Lite, it outperforms the IRRG-only classfier by 5\% out-of-sample. For comparison, Table \ref{isprs_results_svm} lists the performance metrics when using the SVM classifier for the ISPRS datasets. As expected, overall performance drops significantly. \\

\begin{table}[ht]
    \centering
    \caption{SegNet - Classification performance by object type (accuracy only) for ISPRS in-sample (Vaihingen) and out-of-sample (Potsdam) validation using three training cases.}
    \begin{tabular}{|x|*{6}{c|}}
    \hline
    \rowcolor{red!20}
        {\;} & \multicolumn{6}{|c|}{\textbf{SegNet (ISPRS)}} \\
    \hline
    \rowcolor{red!20}
        & \multicolumn{3}{|c|}{\textbf{Vaihingen}} & \multicolumn{3}{|c|}{\textbf{Potsdam (9 cm)}} \\
    \hline
    \rowcolor{red!20}
        \textbf{Objects of Interest} & \textbf{nDSM} & \textbf{IRRG} & \textbf{nDSM} & \textbf{nDSM} & \textbf{IRRG} & \textbf{nDSM} \\
    \rowcolor{red!20}
        &  &  & \textbf{\& IRRG} &  &  & \textbf{\& IRRG} \\
    \hline
        Impervious surfaces & 0.8727 & 0.9520 & \textbf{0.9531} & 0.7127 & 0.7502 & \textbf{0.8374} \\
    \hline
        Buildings & 0.9549 & 0.9738 & \textbf{0.9722} & 0.6828 & 0.4571 & \textbf{0.7886} \\
    \hline
        Low vegetation & 0.8486 & 0.9299 & \textbf{0.9243} & 0.7320 & 0.7829 & \textbf{0.8589} \\
    \hline 
        Trees & 0.9159 & \textbf{0.9488} & 0.9473 & \textbf{0.8846} & 0.8568 & 0.8643 \\
    \hline
        Cars & 0.9922 & \textbf{0.9969} & 0.9959 & 0.9865 & 0.9879 & \textbf{0.9912} \\
    \hline
        Clutter & 0.9995 & 0.9993 & \textbf{0.9996} & 0.9518 & \textbf{0.9598} & 0.9522 \\
    \hline
        Total & 0.7919 & \textbf{0.9003} & 0.8962 & 0.4752 & 0.3974 & \textbf{0.6463} \\
    \hline
    \end{tabular}

    \label{isprs_results_full_segnet}
\end{table}
\begin{table}[ht]
    \centering
    \caption{SegNet Lite - Classification performance by object (accuracy only) for ISPRS in-sample (Vaihingen) and out-of-sample (Potsdam) validation using three training cases.}
    \begin{tabular}{|z|*{6}{c|}}
    \hline
    \rowcolor{cyan}
        {\;} & \multicolumn{6}{|c|}{\textbf{SegNet Lite (ISPRS)}} \\
    \hline
    \rowcolor{cyan}
         & \multicolumn{3}{|c|}{\textbf{Vaihingen}} & \multicolumn{3}{|c|}{\textbf{Potsdam (9 cm)}} \\
    \hline
    \rowcolor{cyan}
        \textbf{Objects of Interest} & \textbf{nDSM} & \textbf{IRRG} & \textbf{nDSM} & \textbf{nDSM} & \textbf{IRRG} & \textbf{nDSM} \\
    \rowcolor{cyan}
        &  &  & \textbf{\& IRRG} &  &  & \textbf{\& IRRG} \\
    \hline
        Impervious surfaces & 0.8706 & 0.9519 & \textbf{0.9559} & 0.7123 & \textbf{0.7950} & 0.7827 \\
    \hline
        Buildings & 0.9539 & 0.9726 & \textbf{0.9735} & \textbf{0.8559} & 0.5554 & 0.6016 \\
    \hline
        Low vegetation & 0.8417 & \textbf{0.9322} & 0.9276 & 0.6077 & 0.7651 & \textbf{0.8182} \\
    \hline 
        Trees & 0.9162 & 0.9490 & \textbf{0.9486} & \textbf{0.8687} & 0.8384 & 0.8669 \\
    \hline
        Cars & 0.9922 & \textbf{0.9969} & 0.9959 & 0.9864 & \textbf{0.9887} & 0.9871 \\
    \hline
        Clutter & 0.9992 & 0.9992 & \textbf{0.9996} & 0.9522 & \textbf{0.9551} & 0.9495 \\
    \hline
        Total & 0.7869 & \textbf{0.9009} & 0.9006 & 0.4916 & 0.4488 & \textbf{0.5030}\\
    \hline
    \end{tabular}
    \label{isprs_results_segnet_lite}
\end{table}

\begin{figure}[ht]
    \centering
    \includegraphics[width=15cm]{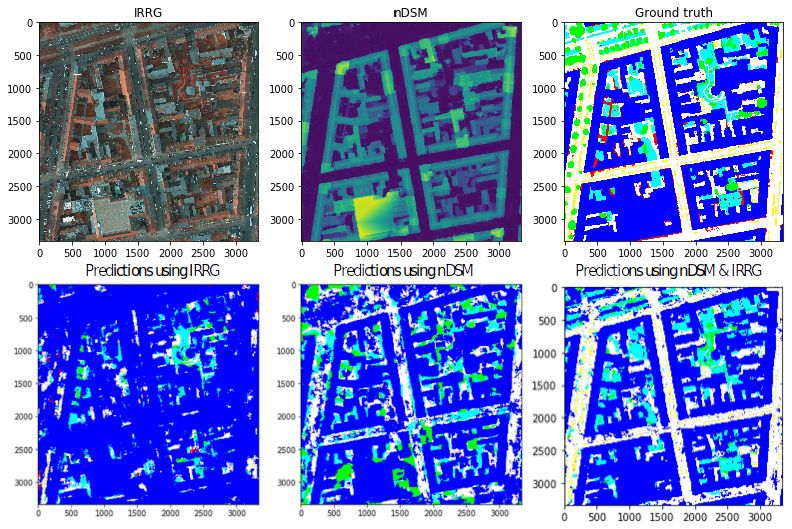}
    \caption{Qualitative out-of-sample classification performance for SegNet classifier applied to ISPRS Potsdam data. From left to right, the top row shows IRRG imagery, nDSM information, color-coded ground truth annotations. From left to right, bottom row display predictions when trained with (i) IRRG info only, (ii) nDSM info only, and (iii) combined IIRG \& nDSM info. }
    \label{segnet_predictions_isprs_fig}
\end{figure}

\begin{figure}[ht]
    \centering
    \includegraphics[width=15cm]{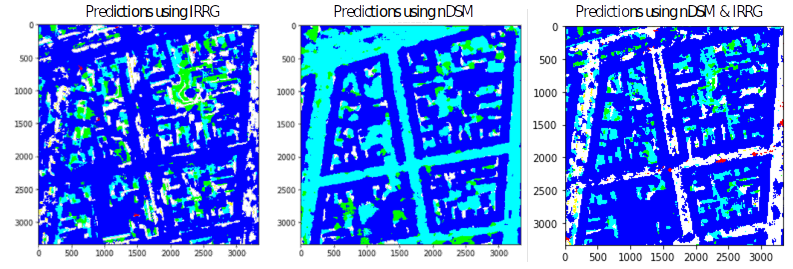}
    \caption{Qualitative out-of-sample classification performance for SegNet Lite classifier for the same tile as used in Figure \ref{segnet_predictions_isprs_fig} from the ISPRS Potsdam data, display predictions when trained with (i) IRRG info only, (ii) nDSM info only, and (iii) combined IIRG \& nDSM info.}
    \label{segnet_lite_predictions_isprs_fig}
\end{figure}

\begin{table}[ht]
    \centering
    \caption{SVM - Classification performance by object (accuracy only) for in-sample (Vaihingen) and out-of-sample (Potsdam) validation using three training cases.}
    \begin{tabular}{|y|*{6}{c|}}
    \hline
    \rowcolor{green!50}
        {\;} & \multicolumn{6}{|c|}{\textbf{5 $\times$ 5 SVM Classifier (ISPRS)}} \\
    \hline
    \rowcolor{green!50}
         & \multicolumn{3}{|c|}{\textbf{Vaihingen}} & \multicolumn{3}{|c|}{\textbf{Potsdam (9 cm)}} \\
    \hline
    \rowcolor{green!50}
        \textbf{Objects of Interest} & \textbf{nDSM} & \textbf{IRRG} & \textbf{nDSM} & \textbf{nDSM} & \textbf{IRRG} & \textbf{nDSM} \\
    \rowcolor{green!50}
        &  &  & \textbf{\& IRRG} &  &  & \textbf{\& IRRG} \\
    \hline
        Impervious surfaces & 0.7812 & 0.8733 & \textbf{0.9320} & 0.6847 & 0.7665 & \textbf{0.8352} \\
    \hline
        Buildings & 0.7931 & 0.8914 & \textbf{0.9567} & \textbf{0.7550} & 0.5257 & 0.6913 \\
    \hline
        Low vegetation & 0.8309 & 0.8715 & \textbf{0.8978} & 0.7246 & 0.7768 & \textbf{0.8147} \\
    \hline 
        Trees & 0.7537 & 0.9101 & \textbf{0.9317} & 0.7464 & \textbf{0.8325} & 0.8214 \\
    \hline
        Cars & 0.9688 & 0.9915 & \textbf{0.9928} & 0.8530 & \textbf{0.9862} & 0.9832 \\
    \hline
        Clutter & 0.9922 & \textbf{0.9997} & \textbf{0.9997} & 0.9412 & \textbf{0.9436} & 0.9429 \\
    \hline
        Total & 0.5600 & 0.7687 & \textbf{0.8553} & 0.3266 & 0.4157 & \textbf{0.5444} \\
    \hline
    \end{tabular}
    \label{isprs_results_svm}
\end{table}
Figure \ref{segnet_predictions_isprs_fig} shows qualitative results for the SegNet architecture when generating predictions using the three training cases. Ground truth is annotated using color-coding for roads (white), buildings (blue), low vegetation (cyan), trees (green), cars (yellow) and clutter (red). Again, without nDSM information, misclassifications occur between buildings, low vegetation, trees and roads. Figure \ref{segnet_lite_predictions_isprs_fig} presents the corresponding qualitative results for the SegNet Lite architecture. \\

\subsubsection{USSOCOM Dataset Results}
\label{us_socom_results}
\begin{figure}[ht]
    \centering
    \includegraphics[width=10cm]{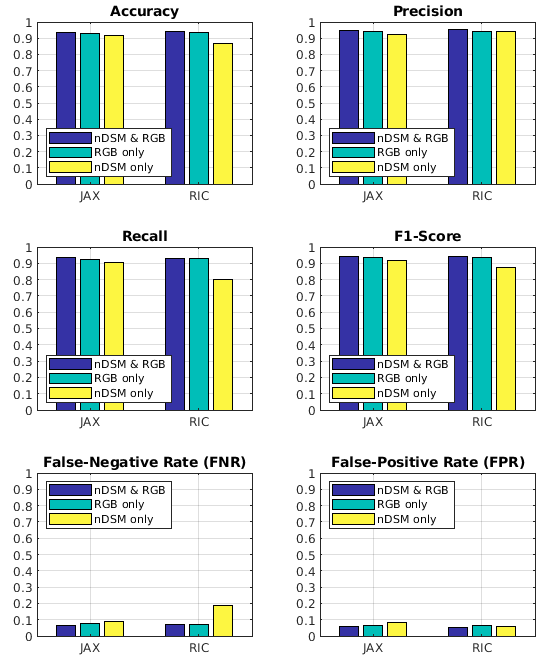}
    \caption{Cross-city building classification performance for the USSOCOM dataset using SegNet classifiers. Classifiers are color-coded: nDSM \& RGB in blue, RGB-only in green, and nDSM-only in yellow. Note that JAX corresponds to out-of-sample testing with tiles from Jacksonville, and RIC corresponds to out-of-sample testing with tiles from Richmond.}
    \label{cross_city_performance_segnet}
\end{figure}

\begin{figure}[ht]
    \centering
    \includegraphics[width=10cm]{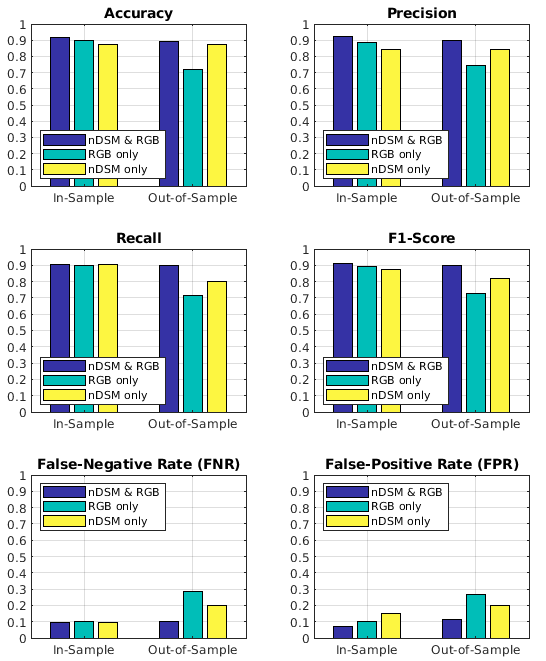}
    \caption{In-sample and out-of-sample building classification performance for the USSOCOM dataset using SVM classifiers. Classifiers are color-coded: nDSM \& RGB in blue, RGB-only in green, and nDSM-only in yellow. }
    \label{cross_city_performance}
\end{figure}

Figures \ref{cross_city_performance_segnet} and \ref{cross_city_performance} summarize the resulting performance for our SegNet and SVM models using quantitative binary classification metrics such as accuracy, precision, recall, F1-score, and false-negative and false-positive rates. Classifiers are color-coded as follows: nDSM \& RGB in blue, RGB-only in green, and nDSM-only in yellow. \\

For SVM (Figure \ref{cross_city_performance}), the left three bars show in-sample performance, i.e. testing was performed on the same 144 tiles that the classifiers were trained on, while the right three bars represent out-of-sample performance, i.e. testing was performed using 144 unseen Tampa tiles. For SegNet (Figure \ref{cross_city_performance_segnet}), the left three bars show out-of-sample performance for procedure 1 (testing on 144 tiles over Jacksonville), while the right three bars represent out-of-sample performance for procedure 2 (testing on 144 tiles over Richmond). \\

Figure \ref{cross_city_performance} indicates that in-sample performance (left most three bars in all six subplots) decreases only slightly when using RGB (green) or nDSM (yellow) information only, as compared to the combined nDSM \& RGB classifier (blue). Note the RGB classifier slightly outperforms the (nDSM) classifier in in-sample performance. However, figures \ref{cross_city_performance_segnet} and \ref{cross_city_performance} indicate that performance differs significantly when testing the trained classifiers on a previously unseen test dataset, here USSOCOM Jacksonville or Richmond tiles (SegNet) and USSOCOM Tampa tiles (SVM). In addition to the overall performance discrepancies between the three classifiers for both SegNet and SVM, it becomes evident that accuracy drops only 10\% when using only nDSM data as compared to 15\% when using only RGB information (SVM; see upper left plot in Figure \ref{cross_city_performance}). For the SegNet classifiers, we observe that classifiers trained with RGB \& nDSM information exhibit an average 0.6\% higher out-of-sample accuracy than classifiers trained on RGB information alone. These results support the hypothesis that the nDSM information facilitates greater classifier generalizability, as compared to RGB information alone for building classification tasks. \\

\begin{figure}[ht]
    \centering
    \includegraphics[width=15cm]{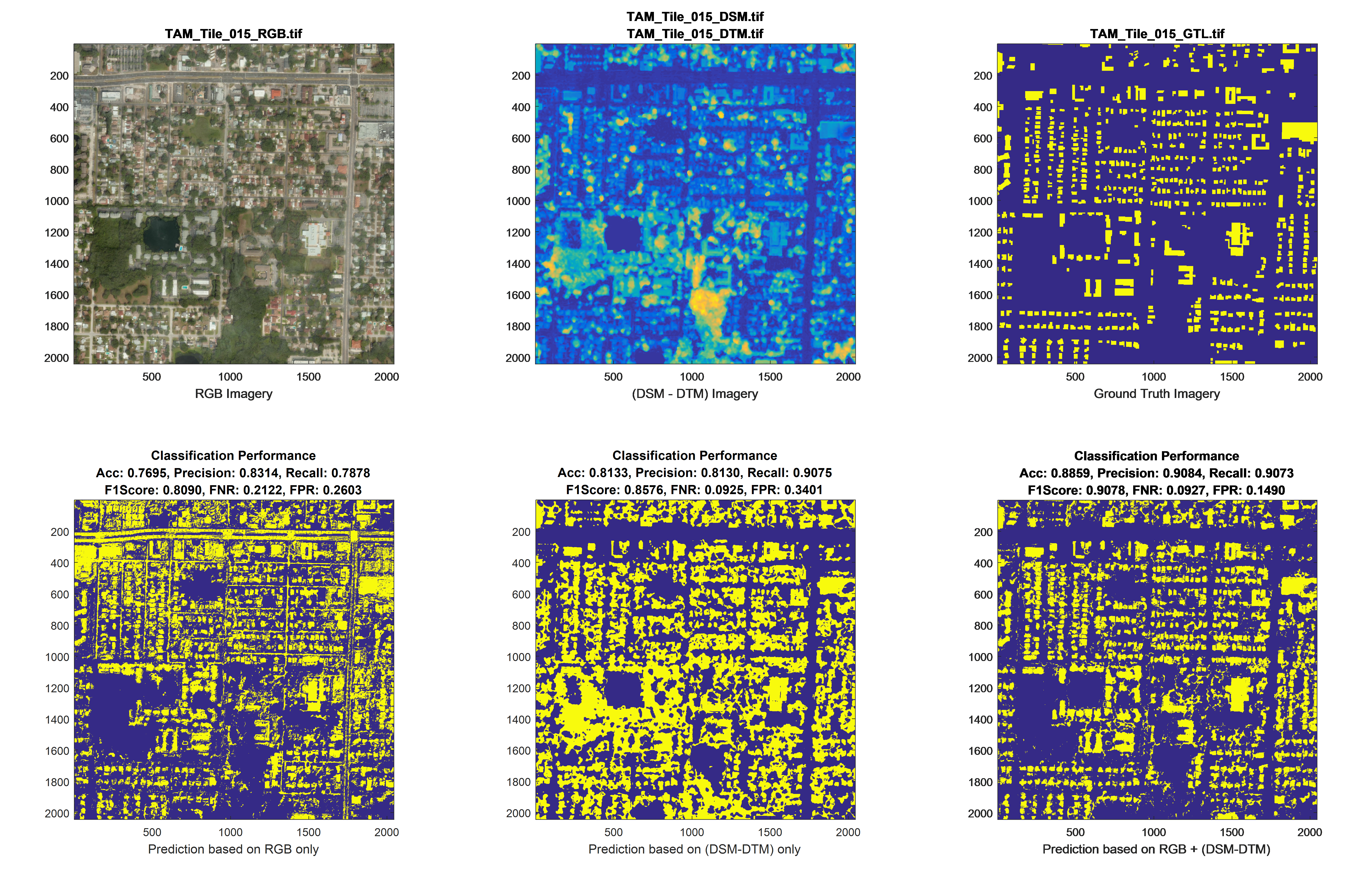}
    \caption{Qualitative out-of-sample classification performance for SVM classifiers applied to USSOCOM data. From left to right, the upper row shows RGB imagery, nDSM (DSM-DTM) information, and ground truth, i.e. annotated building footprints, for Tampa tile \#014. From left to right, the lower row shows predicted building footprints when training on (i) nDSM information only, (ii) RGB imagery only, and (iii) combined RGB \& nDSM information.}
    \label{svm_results}
\end{figure}

Figure \ref{svm_results} presents the qualitative out-of-sample performance for all three SVM classifiers. From left to right, the upper row shows the training data RGB imagery, nDSM information, and ground truth, i.e. annotated building footprints, for Tampa tile \#014. From left to right, the lower row shows predicted building footprints when training on (i) RGB information only, (ii) nDSM information only, and (iii) combined RGB \& nDSM information. It is clear that the RGB \& nDSM classifier on the lower right provides the best correlation with the actual ground truth (upper right). However, specific misclassifications occur for the other two cases. For example, when using nDSM information only, taller non-building objects such as trees are associated with a higher misclassification rate. In contrast, when using RGB information only, objects such as roads are often misclassified as buildings. However, when combining RGB and nDSM information, the number of misclassifications (both Type I and Type II) is significantly reduced. \\

Table \ref{ussocom_results_full_segnet} captures our results from applying SegNet to the USSOCOM dataset with the training procedures specified in Table \ref{ussocom_training}. Similarly, Table \ref{ussocom_results_svm} captures our results from applying our SVM classifier to the same dataset. \\

\begin{table}[ht]
    \centering
    \caption{SegNet - Balanced building classification performance metrics for cross-city (out-of-sample) validation following procedures 1 and 2 in table \ref{ussocom_training}. In procedure 1 (left three columns), SegNet was trained on tiles from Tampa, Florida, and tested on tiles from Jacksonville, Florida. In procedure 2 (right three columns), SegNet was trained on tiles from Tampa, Florida, and tested on tiles from Richmond, Virginia.}
    \begin{tabular}{|x|*{6}{c|}}
    \hline
    \rowcolor{red!20}
         & \multicolumn{6}{|c|}{\textbf{SegNet (USSOCOM)}} \\
    \hline
    \rowcolor{red!20}
         & \multicolumn{3}{|c|}{\textbf{Train TAM}} & \multicolumn{3}{|c|}{\textbf{Train TAM}} \\
    \rowcolor{red!20}
         & \multicolumn{3}{|c|}{\textbf{Test JAX}} & \multicolumn{3}{|c|}{\textbf{Test RIC}} \\
    \hline
    \rowcolor{red!20}
        \textbf{Classification} & \textbf{nDSM} & \textbf{RGB} & \textbf{nDSM} & \textbf{nDSM} & \textbf{RGB} & \textbf{nDSM} \\
    \rowcolor{red!20}
       \textbf{Metrics} &  &  & \textbf{\& RGB} &  &  & \textbf{\& RGB} \\
    \hline
        Accuracy & 0.9164 & 0.9298 & \textbf{0.9367} & 0.8690 & 0.9339 & \textbf{0.9386} \\
    \hline
        Precision & 0.9245 & 0.9412 & \textbf{0.9451} & 0.9425 & 0.9416 & \textbf{0.9512} \\
    \hline
        Recall & 0.9105 & 0.9245 & \textbf{0.9341} & 0.8122 & \textbf{0.9307} & 0.9298 \\
    \hline
        F1 Score & 0.9175 & 0.9328 & \textbf{0.9396} & 0.8725 & 0.9361 & \textbf{0.9404} \\
    \hline
        False Negative Rate & 0.0895 & 0.0755 & \textbf{0.0659} & 0.1878 & \textbf{0.0693} & 0.0702 \\
    \hline
        False Positive Rate & 0.0829 & 0.0643 & \textbf{0.0604} & 0.0610 & 0.0626 & \textbf{0.0518} \\
    \hline
    \end{tabular}
    \label{ussocom_results_full_segnet}
\end{table}

\begin{table}[ht]
    \centering
    \caption{SVM - Balanced building classification performance metrics for in-sample and out-of-sample testing on the USSOCOM dataset.}
    \begin{tabular}{|y|*{6}{c|}}
    \hline
    \rowcolor{green!50}
          & \multicolumn{6}{|c|}{\textbf{5 $\times$ 5 SVM Classifier (USSOCOM)}} \\
    \hline
    \rowcolor{green!50}
         & \multicolumn{3}{|c|}{\textbf{In-sample}} & \multicolumn{3}{|c|}{\textbf{Out-of-sample}} \\
    \rowcolor{green!50}
         & \multicolumn{3}{|c|}{\textbf{Testing}} & \multicolumn{3}{|c|}{\textbf{Testing}} \\
    \hline
    \rowcolor{green!50}
        \textbf{Classification} & \textbf{nDSM} & \textbf{RGB} & \textbf{nDSM} & \textbf{nDSM} & \textbf{RGB} & \textbf{nDSM} \\
    \rowcolor{green!50}
       \textbf{Metrics} &  &  & \textbf{\& RGB} &  &  & \textbf{\& RGB} \\
    \hline
        Accuracy & 0.8763 & 0.8978 & \textbf{0.9178} & 0.8763 & 0.7212 & \textbf{0.8931} \\
    \hline
        Precision & 0.8438 & 0.8850 & \textbf{0.9214} & 0.8438 & 0.7467 & \textbf{0.9003} \\
    \hline
        Recall & \textbf{0.9047} & 0.8996 & 0.9023 & 0.8000 & 0.7126 & \textbf{0.8963} \\
    \hline
        F1 Score & 0.8732 & 0.8922 & \textbf{0.9117} & 0.8200 & 0.7292 & \textbf{0.8983} \\
    \hline
        False Negative Rate & \textbf{0.0953} & 0.1004 & 0.0977 & 0.2000 & 0.2874 & \textbf{0.1037} \\
    \hline
        False Positive Rate & 0.1488 & 0.1039 & \textbf{0.0684} & 0.2000 & 0.2693 & \textbf{0.1105} \\
    \hline
    \end{tabular}
    \label{ussocom_results_svm}
\end{table}
\subsection{Validation Using Small Sample Proportion}
\label{small_sample_validation}
In this section, the importance of 3D surface information is further tested using classification scenarios with scarce training samples. Sufficient data with adequate representation and viewing angles for all objects of interest may not always be assumed, particularly for remote sensing applications. Therefore, we train and test the two classification architectures from Section \ref{ClassificationArchitectures}, while successively decreasing the number of training samples. \\

SVM classification in Section \ref{cross_city_validation} was carried out using all 144 Jacksonville tiles from the ISPRS dataset. The 600 million training samples, i.e. annotated pixels, were randomly down-selected by a factor of 1,000 to 600,000 training samples, which corresponds to a sample proportion for training of 0.1\%. For the following analysis, we further decrease the sample proportion to 0.01\%, 0.001\% and 0.0001\%, thereby reducing the total number of training samples to 60,000, 6,000, and 600, respectively. \\

Table \ref{fraction_training} presents the resulting average training times for all three SVM classifiers, highlighting the orders of magnitude between the different test cases. Clearly, the underlying numerical optimization can be completed significantly faster if fewer training samples need to be classified. \\

\begin{table}[ht]
    \centering
    \caption{Average training times (in seconds) for SVM classifiers when using smaller sample proportions for training on USSOCOM data.}
    \label{fraction_training}
    \begin{tabular}{|a|c|c|c|c|}
        \hline
        Sample Proportion for Training & 0.0001\% & 0.001\% & 0.01\% & 0.1\% \\
        \hline
        SVM Training Times (sec) & 0.1 & 1.5 & 180 & 24,000 \\
        \hline
    \end{tabular}
\end{table}

\begin{figure}[ht]
    \centering
    \includegraphics[width=6.0in]{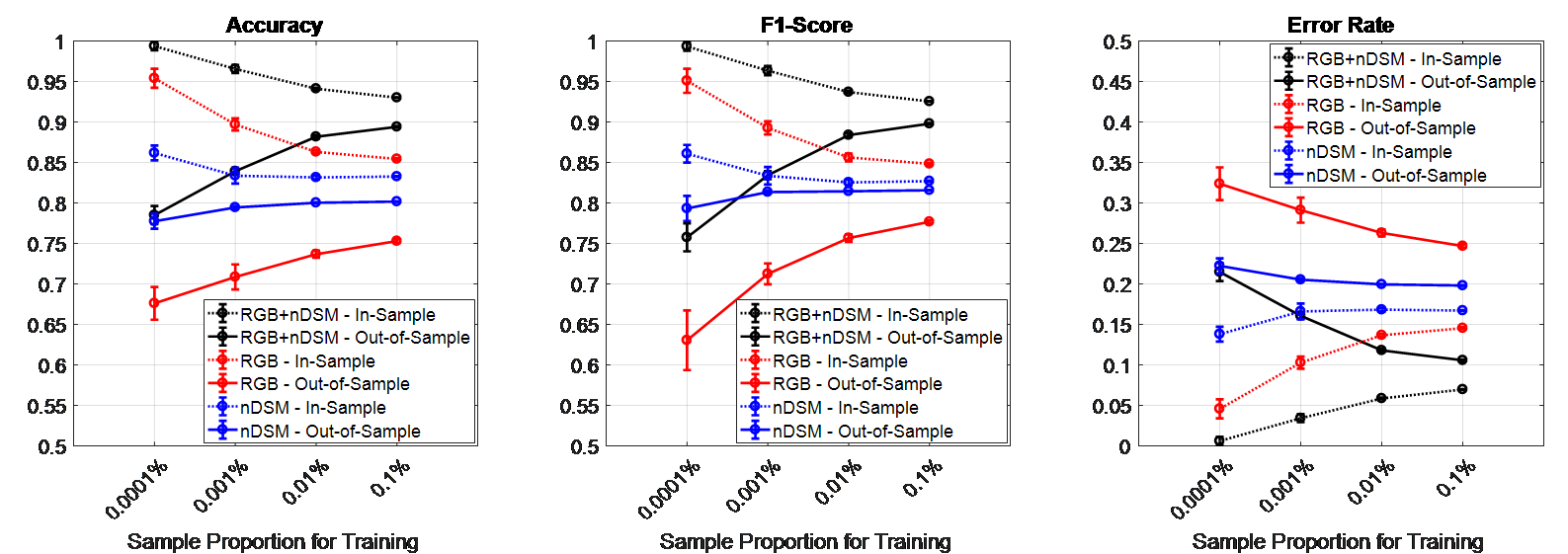}
    \caption{Impact of sample proportion on in-sample (dotted lines) and out-of-sample (solid lines) SVM classification performance on the USSOCOM Jacksonville, FL dataset. The study compares three input data scenarios, (a) RGB \& nDSM (black), (b) RGB-only (red), and (c) nDSM-only (blue). From left to right, the individual plots show accuracy, F1-score, and error rate as a function of sample proportion.}
    \label{sample_prop_vs_performance_fig}
\end{figure}

Figure \ref{sample_prop_vs_performance_fig} displays the resulting in-sample and out-of-sample classification performance for the three SVM classifiers: RGB-only (red), nDSM-only (blue), and RGB \& nDSM (black) as a function of sample proportion for training. Here, performance is measured in accuracy (left plot), F1-score (center plot) and error rate (right plot). All metrics assume class balancing. In-sample performance is plotted as dotted lines, while out-of-sample performance is plotted as solid lines. As training and test samples are selected randomly, five trials were conducted for each test case studied. In Figure \ref{sample_prop_vs_performance_fig}, vertical bars are added to indicate the standard deviation for the particular performance metric over those five trials. \\
As discussed in the previous section, the RGB \& nDSM classifier provides the best in-sample performance at 93\% accuracy when using 0.1\% of all training data. In-sample performance for the RGB-only and nDSM-only classifiers is 85\% and 83\%, respectively. In-sample accuracy (dotted lines) increases for all three classifiers as the sample proportion for training decreases. This is due to the fact that fewer training samples have to be classified. However, out-of-sample performance for all classifiers decreases with decreasing sample proportion for training, indicating that the resulting classifiers lose their generalizability when trained on smaller training sets due to overfitting. For out-of-sample performance, the nDSM-only classifier outperforms the RGB-only classifier, which further affirms the findings from Section \ref{cross_city_validation}. Interestingly, nDSM-only even outperforms the RGB \& nDSM in the 0.0001\% case. This result may relate to the curse of dimensionality \bracketcite{Keogh2017}, as the nDSM classifier operates in a reduced feature space of 25 dimensions (see Section \ref{svm_background}), while the combined RGB \& nDSM classifier operates in 100 dimensions. In general, if training data is scarce, a reduced feature space can improve generalizability by avoiding overfitting. \\

\begin{figure}[ht]
  \centering
  \includegraphics[width=15cm]{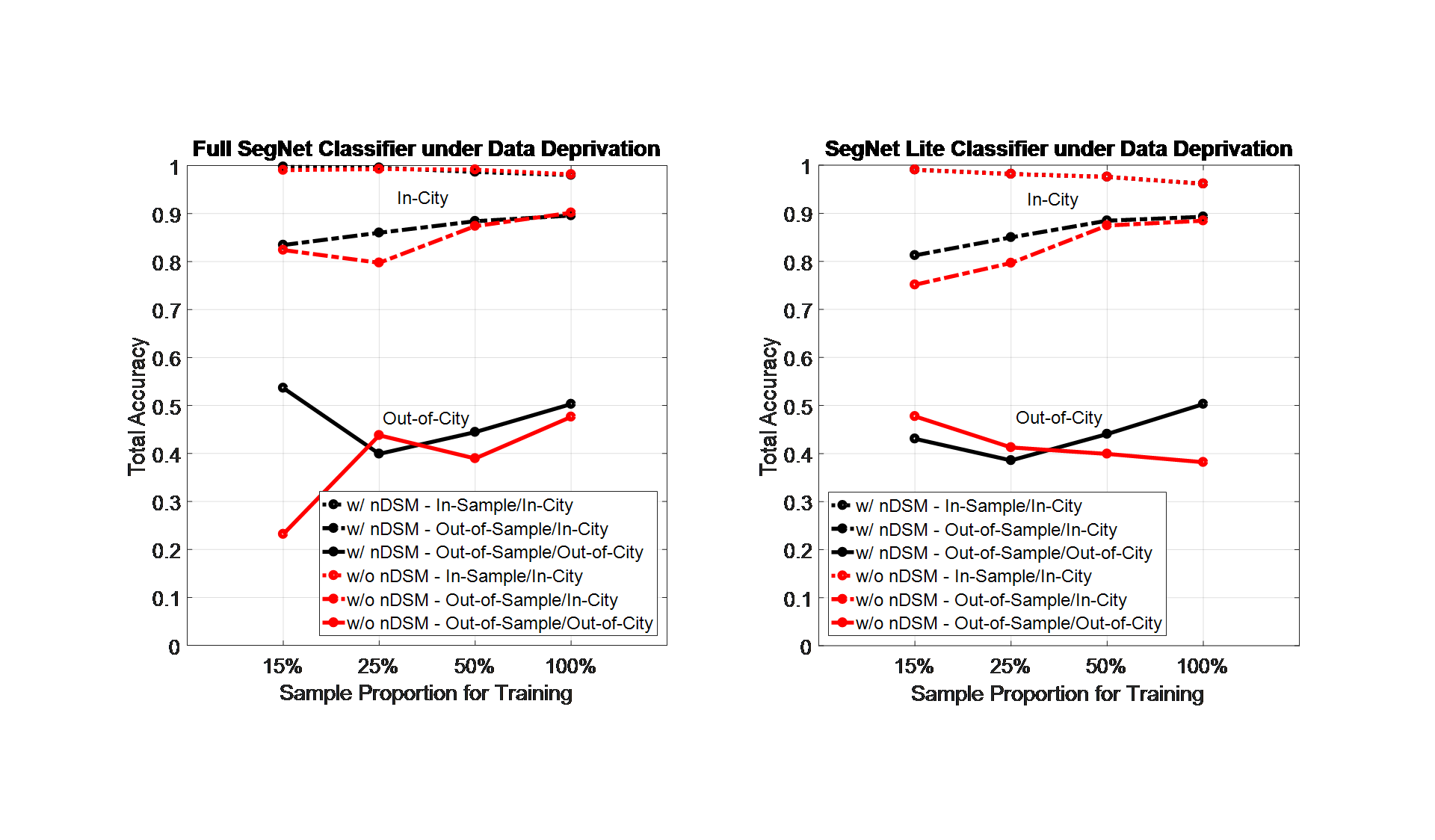}
  \caption{Impact of sample proportion on classification performance using SegNet (left) and SegNet Lite (right) on ISPRS data.}
  \label{isprs_impact}
\end{figure}

In addition to the SVM classifiers, we conduct the same validation analysis for small sample proportion for training the two SegNet architectures from Section \ref{segnet_background}. Figure \ref{isprs_impact} displays the results when training the SegNet and SegNet Lite classifiers with 15\%, 25\%, 50\% and 100\% of the data. The method used for obtaining a subset of the data is to select a random point from each image and take a width and height equal to the desired fraction of the full image as the cropping region, which is then used for training. As before, training was carried out using 12 ISPRS Vaihingen tiles. Testing was then performed for three cases: (i) in-sample/in-city (using the 12 Vaihingen tiles that were used for training), (ii) out-of-sample/in-city (using the remaining 4 Vaihingen tiles not used from training), and (iii) out-of-sample/out-of-city (using all ISPRS Potsdam tiles).
Out-of-sample/cross-city accuracy across the SegNet and Segnet Lite models with and without nDSM generally indicate a mild positive correlation between portion of data and accuracy, suggesting that 50\% of the data for a given city might be sufficient for the full city classification. \\

In-sample/in-city accuracy across the SegNet and Segnet Lite models with and without nDSM exhibits a negative correlation between portion of dataset and accuracy. As with the SVM classifier, this can be attributed to the network having less samples to classify, and therefore being able to overfit to the scarce training set. Lastly, the non-nDSM trained SegNet model has a negative correlation between accuracy and training proportion in regards to cross-city testing. This correlation may indicate that, in addition to the cross-validation results of RGB and nDSM, RGB information alone can lead to overfitting and therefore hinder generalizability of a classification model to other cities. 
\section{Conclusion}
\label{conclusion}
This paper evaluated the importance of 3D surface information for remote sensing using several classification tasks. Two distinct classifiers, i.e. SVM and FCNN architectures, were introduced and assessed for performance when trained with and without nDSM information. Two publicly available datasets, i.e. the USSOCOM Urban 3D challenge \bracketcite{Urban3D2017}\bracketcite{goldberg2018urban} and the ISPRS 2D Semantic Labeling contests \bracketcite{inproceedings}, were utilized for training and rigorous in-sample and out-of-sample performance assessment. In all cases, the study demonstrated that high in-sample classification performance can be maintained even when depriving the classifier of nDSM information. However, out-of-sample performance, i.e. when testing the classifier on previously unseen data from a different city, drops significantly for both SVM and FCNN classifiers trained without nDSM information. We conclude that nDSM information is vital for accurately generalizing classification methods to datasets not included in training. \\

An additional study revealed that nDSM information is also critical when training a classifier with relatively few training samples. Again, in-sample performance remains high with and without nDSM information, but generalizability decreases substantially when nDSM information is excluded from training. \\

Together, these validation experiments demonstrate the importance of including nDSM information to ensure generalizable out-of-sample predictive performance for remote sensing classification tasks.
\section{Acknowledgements}
\label{acknowledgement}
This work has been supported by the National Geospatial-Intelligence Agency under contract No. N0024-12-D-6404 DO 18F8332, as well as through the Science, Mathematics, and Research for Transformation (SMART) program. We would like to thank Kateri Garcia, Chris Clasen, and Patrick Yuen for their excellent guidance and input on this research. Approved for public release, 21-153. \\

We would also like to thank USSOCOM as well as the ISPRS for providing public datasets including annotations of objects of interest to facilitate research in remote sensing and machine learning. In addition, we thank the authors of \bracketcite{audebert2018beyond} for providing the source code for the SegNet architecture via GitHub. The corresponding author, Jan Petrich, may be reached at \href{mailto:jup37@arl.psu.edu}{jup37@arl.psu.edu}.
\section{Competing Interests}
The authors declare that they have no competing interests.
\theendnotes
\printbibliography
\section{Appendix: Figure and Table Lists and Legends}
\listoffigures
\listoftables
\end{document}